\documentclass[10pt,twocolumn,letterpaper]{article}

\usepackage{cvpr}              

\usepackage{graphicx}
\usepackage{amsmath}
\usepackage{amssymb}
\usepackage{booktabs}

\newcommand*\samethanks[1][\value{footnote}]{\footnotemark[#1]}

\usepackage[pagebackref,breaklinks,colorlinks]{hyperref}

\usepackage[capitalize]{cleveref}
\crefname{section}{Sec.}{Secs.}
\Crefname{section}{Section}{Sections}
\Crefname{table}{Table}{Tables}
\crefname{table}{Tab.}{Tabs.}

\def\cvprPaperID{*****} 
\def\confName{CVPR}
\def\confYear{2022}

\begin{document}

\title{MAE-GEBD:Winning the CVPR'2023 LOVEU-GEBD Challenge}

\author{\hspace{7mm} Yuanxi Sun                        \textsuperscript{1} \hspace{14mm} Rui He                \textsuperscript{1} \hspace{10mm} Youzeng Li            \textsuperscript{1} \hspace{10mm} Zuwei Huang           \textsuperscript{1}\\
Feng Hu\textsuperscript{ 1}  \hspace{17mm} Xu Cheng\thanks{Corresponding Authors.}\ \ \textsuperscript{1 2} \hspace{14mm} Jie Tang\samethanks \ \ \textsuperscript{2}\\
Tencent Holdings Ltd.\textsuperscript{1}\\
Department of Computer Science and Technology, Tsinghua University\textsuperscript{ 2}\\
{\tt\small \{yuanxisun,rayruihe,youzengli,takumihuang,emonhu,alexcheng\}@tencent.com}\\
\tt\small jietang@tsinghua.edu.cn}

\maketitle

\begin{abstract}
    The Generic Event Boundary Detection (GEBD) task aims to build a model for segmenting videos into segments by detecting general event boundaries applicable to various classes. In this paper, based on last year's MAE-GEBD method, we have improved our model performance on the GEBD task by adjusting the data processing strategy and loss function. Based on last year's approach, we extended the application of pseudo-label to a larger dataset and made many experimental attempts. In addition, we applied focal loss to concentrate more on difficult samples and improved our model performance. Finally, we improved the segmentation alignment strategy used last year, and dynamically adjusted the segmentation alignment method according to the boundary density and duration of the video, so that our model can be more flexible and fully applicable in different situations. With our method, we achieve an F1 score of 86.03\% on the Kinetics-GEBD test set, which is a 0.09\% improvement in the F1 score compared to our 2022 Kinetics-GEBD method. Our code is available at \
    \href{https://github.com/ContentAndMaterialPortrait/MAE-GEBD}{\color{blue}{https://github.com/ContentAndMaterialPortrait/MAE-GEBD}}.
\end{abstract}

\section{Introduction}
\label{sec:intro}

\begin{figure*}
  \centering
  \includegraphics[width=0.8\linewidth]{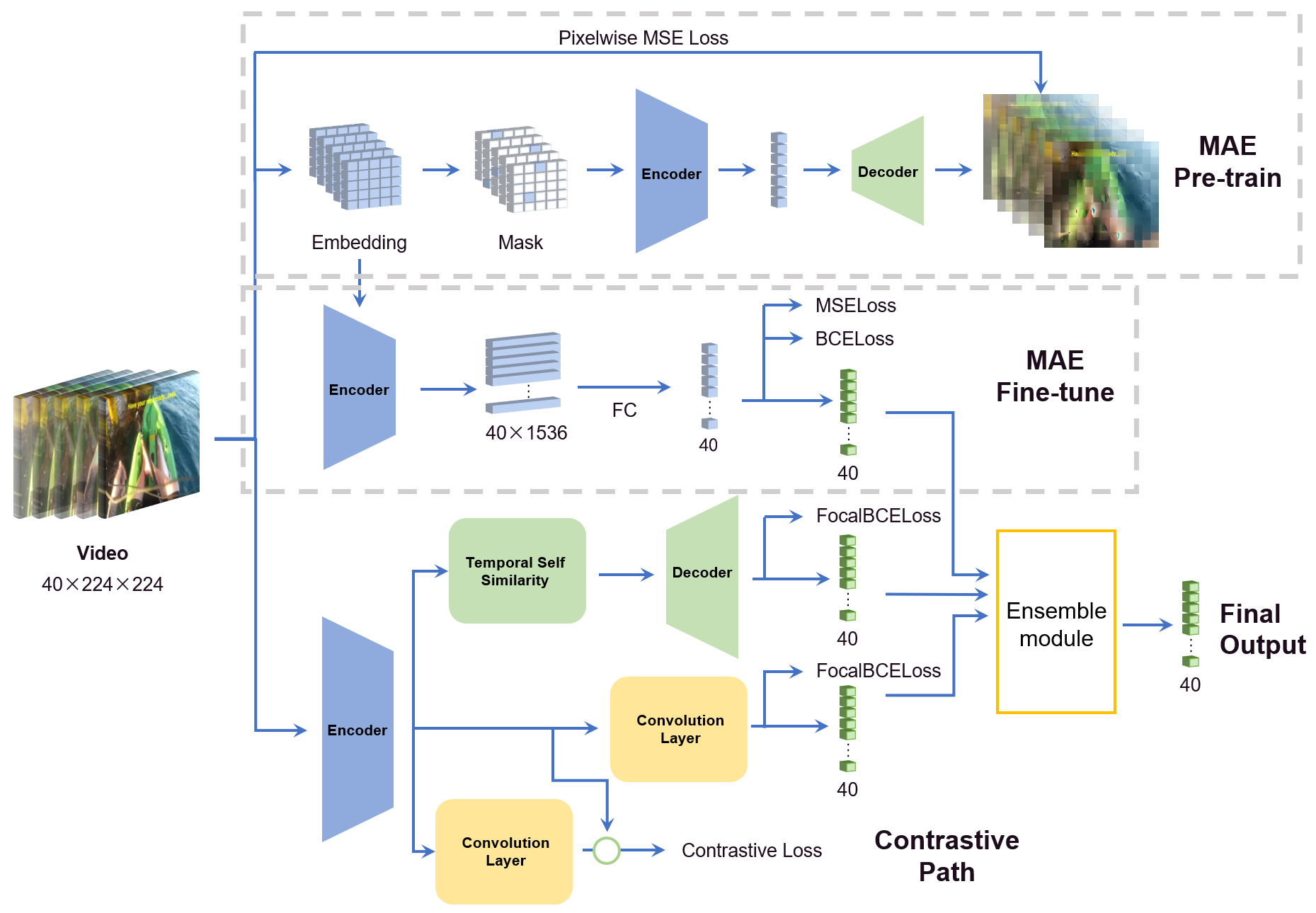}
  \caption{MAE-GEBD model structure.}
   \label{fig:model}
\end{figure*}

Vision is the most important way for human beings to obtain information. When people receive continuous visual signals through their eyes and brains, people tend to divide them into different segments according to the content of the video signal, so as to help the human brain to better understand, simply and process the information. Such a process may be very natural to the human brain, but not to the machine. In order for a model to mimic this information-gathering mechanism, the Generic Event Boundary Detection (GEBD) task was proposed to localize the boundaries automatically, around which the event content changes\cite{DBLP:journals/corr/abs-2101-10511}.

The current mainstream model \cite{DBLP:journals/corr/abs-2101-10511,jwkim,DBLP:journals/corr/abs-2107-00239}  converts the GEBD task into a binary classification task. The model first slices the video into short segments of equal length, and each segment is considered as a positive sample if there is a boundary in the segment, and a negative otherwise. Our method last year \cite{he2022masked} basically followed this scheme, and adopted a Masked Autoencoder model fine-tuned on the GEBD task to improve the result. By weighting the labeling results and dynamically adjusting the boundary position, the prediction results got rid of the restriction of the fixed-length short segments and became more flexible and accurate.

Since the definition of an event boundary is subjective, we proposed Pseudo-Label \cite{lee2013pseudo} last year to solve the problem of inconsistent results marked by different people. We tried to use the same method this time to obtain more training data without manual labeling in Kinetics-400k, and use these pre-labeled results in the model training.

Based on last year's experience, the number of cuts of video data varies greatly, so it may be difficult to find one cutting point for some videos, while other videos may be cut into more than 10 segments. In order to make the model solve those cases with more boundaries easily, we added the focal loss to the model loss function to make the model care more about those difficult problems.

In the previous scheme, we found that the method of the model predicting whether to cut or not through equal-length video clips will cause the prediction results to only appear in specific positions. To this end, we propose a weighted adjustment based on the prediction score and a segmentation alignment method based on the data distribution. In this project, we explored the relationship between the number of slices and the prediction accuracy of the model. We dynamically adjusted the alignment method according to these features, to make the model have better performance in many cases.

\section{Related Work}
\label{sec:related}

\subsection{Generic Event Boundary Detection}
In \cite{jwkim}, the contrastive learning method won the Kinetics-GEBD challenge in 2021. They proposed the use of temporal self-similarity matrices (TSM) and contrastive loss to solve the problem of event boundaries. In addition to the contrastive learning approach, they also applied the direct prediction approach without TSM or contrastive loss. Those different branches are combined to form the entire model to make a final prediction. In the Kinetics-GEBD challenge in 2022, we adopted the same network structure and proposed some improvement strategies to optimize the performance of this model \cite{he2022masked}. Therefore, we still used the contrastive learning and direct prediction approaches as a part of our method and improved the model performance further. 

\subsection{Masked Autoencoders}
Masked Autoencoders is a popular self-supervised method proposed by \cite{MAE} and gained state-of-the-art performance on various downstream tasks in 2D image domain. Inspired by the MAE, VideoMAE\cite{ VideoMAE} extended masked autoencoder from the 2D image field to the 3D video field and proposed customized video tube masking and reconstruction, and show that video masked autoencoders are data-efficient learners for self-supervised video pre-training (SSVP). Spatiotemporal MAE \cite{MAESpatiotemporal} also proved that the general framework of masked autoencoding (BERT, MAE, etc.) can be a unified methodology for representation learning with minimal domain knowledge. \cite{he2022masked} applied the MAE module on the generic event boundary detection(GEBD) task and won the second place in the Kinetics-GEBD challenge 2022.

\section{Proposed Method}
\label{sec:method}

\subsection{Masked Autoencoders}
Same as in \cite{he2022masked}, we applied MAE in our models dealing with the GEBD task in two stages. In the pre-training stage, for a $40\times224\times224$ video, we got $20\times14\times14$ patches with a patch size of $2\times16\times16$. Referring to Spatiotemporal MAE \cite{MAESpatiotemporal}, the mask ratio is 90\%. We use a pixel-wise reconstruction loss to pre-train our model and get the video embedding and encoder. In the fine-tuning stage, we designed two headers using the BCELoss and MSELoss respectively to predict which frame the boundaries are located in. The embedding and encoder parameters at the fine-tuning stage are inherited from the end of the pre-training stage, and all these parameters would be changed in the fine-tuning stage. 

\subsection{Pseudo-Label}
\label{Pseudo-Label-methods}
In our last year's method \cite{he2022masked}, the pseudo-label strategy improved model performance both on validation and test set, whether it is using the MAE method or the contrastive learning method. Therefore, we still use pseudo-label in our method this year. Same as in \cite{he2022masked}, we trained a relatively weak model using all labeled training and validation sets, and use the weak model to label the unlabeled test set data. We called the annotations of the test set gained from the weak model "pseudo-labels". Then, We use all ground-truth of the training and validation sets as well as all pseudo-labels of the test set to train stronger models.

Moreover, we pseudo-labeled more data from Kinetics-400 which are not in the Kinetics-GEBD dataset this year, we hope more pseudo-labels could further improve model performance. However, these pseudo-labels of non-Kinetics-GEBD dataset didn't improve model performance. Conversely, they are detrimental to model performance. The detailed experimental result can be seen as \cref{experiment:pseudolabel}

\subsection{Focal Loss}
We consider that the difficulty of the GEBD task is highly correlated with the number of boundaries of each video. More boundaries means that the content of this video is more complex and it is more difficult to detect where the boundaries are located. For this reason, on BCE loss, we applied the focal loss strategy, We assign higher loss weights to videos with more boundaries in the annotation, and the loss is computed as follows:
$$\text{focal loss} = \sum_{i=1}^{S}\frac{n[i]}{10}{\text{ BCE loss}[i]}$$
where $S$ is the number of training samples, $n[i]$ is the number of boundaries in the $i\text{-th}$ sample.
By using the focal loss strategy, our model gained a 0.05\% improvement on the F1 score.

\subsection{Segmentation Alignment}

According to the distribution of this data set, we developed a post-processing strategy called segmentation alignment in last year's competition to make our prediction results more efficient. To put it simply, since manual annotations will not appear within 0.3 seconds of the beginning and end of the video, and the range of $[5\% \times \text{duration}]$ before and after each prediction result is regarded as the true positive area, our strategy will shift according to these conditions to make the true positive area cover a larger range. You may refer to \cite{he2022masked} for more detail.

In this year's new strategy, since every two predictions will be separated by 10\% of the video length, there will be generally no more than 10 predictions in each video, which may lead to worse results. Thus we decided to dynamically adjust the segmentation alignment strategy based on both the length of the video and the number of predictions made by the model. For example, if there are more than 10 segmentation positions predicted, we would allow the interval between two split spots to be appropriately reduced to less than $[5\% \times \text{duration}]$. The test results show that the new version of the segmentation alignment strategy can bring about a 0.03\% increase in F1 score in the validation set, and this optimization is basically the same on the test set.

\section{Experiment}
\label{sec:experiment}

\subsection{Dataset and Feature}

Same as in \cite{he2022masked}, we mixed both Kinetic-GEBD original training set (\textsuperscript{\texttildelow}18k samples) and validation set (\textsuperscript{\texttildelow}18k samples), and random split these samples into 10 folds. During the training process, nine of these 10 folds' ground-truth along with all test set pseudo labels (totally \textsuperscript{\texttildelow}50k samples) are used as our training set, and the remaining 1 fold data are used as our validation set. We repeat this training process 10 times while changing our validation set by turns. 

Our MAE-based models' features are gained from the MAE fine-tuning stage. Our contrastive learning models' input features are Kinetics pre-trained two-stream TSN features and SlowFast features, which are the same as \cite{jwkim}.



\subsection{Easy-Hard sample splits}
Same as in \cite{he2022masked}, when training our MAE-based models, we split all samples into two subsets. At the begining, we trained three basic MAE-based models. Using these basic models, we detected boundaries on all samples including those in the test set, and got a score curve for each video sample. If the score curve is flat, we consider this video as a hard sample, whereas if the curve is bumpy, we consider it an easy sample. Then, We trained two different models with easy samples and hard samples respectively. Finally, we use the easy-sample model to predict the easy samples, and the hard-sample model to predict the hard samples in the test set.

Experiments show that both the easy-sample model and the hard-sample model can steadily improve performance when only ensemble MAE-based models. However, when ensemble both MAE-based models and contrastive learning models, only the easy-sample MAE-based model can improve the overall F1-score further, while the hard-sample model was not helpful.

\begin{table}
  \centering
  \begin{tabular}{lcc}
    \toprule
    \textbf{Method}        & \textbf{Test F1-Score} & \textbf{Improvement} \\
    \midrule
    Baseline\cite{jwkim}   & 83.63             &                      \\
    MAE-GEBD 2022\cite{he2022masked}   & 85.94    & + 2.31          \\
    + Segment alignment        & 85.98             & + 0.04         \\
    + Focal Loss        & 86.03             & + 0.05         \\
    \midrule
    \textbf{Test Score}    & 86.03             & + 2.40             \\
    \bottomrule
  \end{tabular}
  \caption{Improvements of each method}
  \label{tab:example}
\end{table}

\subsection{Ensemble}
Same as in \cite{he2022masked}, we trained three kinds of models using different structures, MAE-based models, contrastive learning models, and direct predict models. For MAE-based models, we trained 2 models with pseudo-labels, and 1 model without such strategy. As we explained in $\text{4.2}$, we also trained 1 hard-sample model and 1 easy-sample model. For contrastive learning models and direct predict models, as we showed in $\text{4.1}$, we split all data into 10 folds, and get 20 models in total by two different network structures. Overall, we get 25 models, and 25 predicted scores for each sample in the test set. Then we weighted sum the 24 of them except the MAE-based model trained only on hard samples. The weight of each contrastive learning and direct predict model is 0.0385, and the weight of each MAE-based model is 0.0575.

\subsection{Pseudo-Label}
\label{experiment:pseudolabel}
As shown in \cref{Pseudo-Label-methods}, we used a pseudo-label strategy in training all models. We extracted features of (\textsuperscript{\texttildelow}28k) non-Kinetics-GEBD videos and acquired their pseudo-labels. We explored the effectiveness of this strategy in relation to the amount and scope of pseudo-labels we used during training. The experiment result is shown as \cref{tab:pseudolabel}.

\begin{table}
  \centering
  \begin{tabular}{lcc}
    \toprule
    \textbf{pseudo-label data scope} & \textbf{val F1-Score} & \textbf{Improvement} \\
    \midrule
    No pseudo-label                  & 82.75                 &                      \\
    5k GEBD test set                 & 83.28                 & + 0.53               \\
    10k GEBD test set                & 83.46                 & + 0.71               \\
    14k GEBD test set                & 83.67                 & + 0.92               \\
    18k GEBD test set                & 83.63                 & + 0.88               \\
    10k non-GEBD data                & 82.78                 & + 0.03               \\
    28k non-GEBD data                & 84.46                 & - 0.29               \\
    \bottomrule
  \end{tabular}
  \caption{Improvements of pseudo-labels}
  \label{tab:pseudolabel}
\end{table}

As we can see, using more pseudo-labels of the GEBD test set generally yields more improvement on F1-Score on the validation set. While using pseudo-labels of non-GEBD data is generally harmful to model performance. According to the experiment result, we trained all final models by using only pseudo labels of the 18k GEBD test set.

\subsection{Segmentation Alignment}
First of all, we further explored the distribution of the data, and computed the distribution of both the ground truth and the model-predicted number of splits. The result is as follows \cref{tab:alignment}:

\begin{table}[h!]
\begin{tabular}{c|llll}
\toprule
             & \textbf{no split} & \textbf{0-0.55/s} & \textbf{0.55-1/s} & \textbf{\textgreater{}1/s} \\ \hline
ground truth & 0.28\%   & 62.13\%  & 36.52\%  & 1.06\%            \\
prediction   & 1.76\%   & 57.99\%  & 40.02\%  & 0.23\%           
\end{tabular}
\caption{The distribution of videos according to the number of splits per second. e.g. 0.55-1/s means that the video will have 0.55 to 1 segment on average every second.}
\label{tab:alignment}
\end{table}

The table above shows the distribution of the number of videos in each segmentation count range. It shows that our model tends to predict results with more segments than the ground truth. Specifically, our model will make more videos that get cut more than 0.55 times per second than the ground truth, whereas the ground truth contains more videos with fewer cuts. Besides, our model tends to predict more videos with no segmentation and fewer videos with more than 1 slice per second, compared with the ground truth.

Therefore, in the experiment, we will roughly classify all the data into 23 categories:

$\bullet$ All videos are divided into four categories according to their duration: 0-4 seconds; 4-8 seconds; 8-10 seconds; about 10 seconds.

$\bullet$ For the first three duration categories, the video can be divided into four categories according to the number of segments: 0 segments; 0-0.55 segments per second; 0.55-1 segment per second; more than 1 segment per second.

$\bullet$ For videos of about 10 seconds, due to the large volume, we divide them into 11 categories according to the number of splits from 0 to 9 and 10+ times instead of splits per second.

Using the old version of the segmentation alignment strategy, there are 10.61\% of the videos get a higher F1 score, and only 2.75\% of the videos get a lower score. We can get more information if we further look into each group. The segmentation alignment strategy has a greater impact on videos with more annotations. This is because only when the number of annotations is sufficient will it be affected by an interval of not less than 1 second, and a "chain push" will occur. Therefore, the shift strategy has the greatest impact on the group of videos such as those that have 10s duration and 10+ times of cuts. The promotion ratio for the group was 29.17\%, and the reduction ratio was 22.57\%.

Compare to the overall situation, the reduction ratio is significantly higher in this group. This may be because our push strategy will cause some videos to be "overcrowded" due to too many predicted splits, and the alignment strategy will directly discard some splits to meet our spacing requirements. This overly aggressive strategy may hurt some of the original correct answers, resulting in overcorrection. Therefore, we consider this group (10s, 10+ splits) as a possible improvement direction.

\begin{figure}[t]
  \centering
   \includegraphics[width=\linewidth]{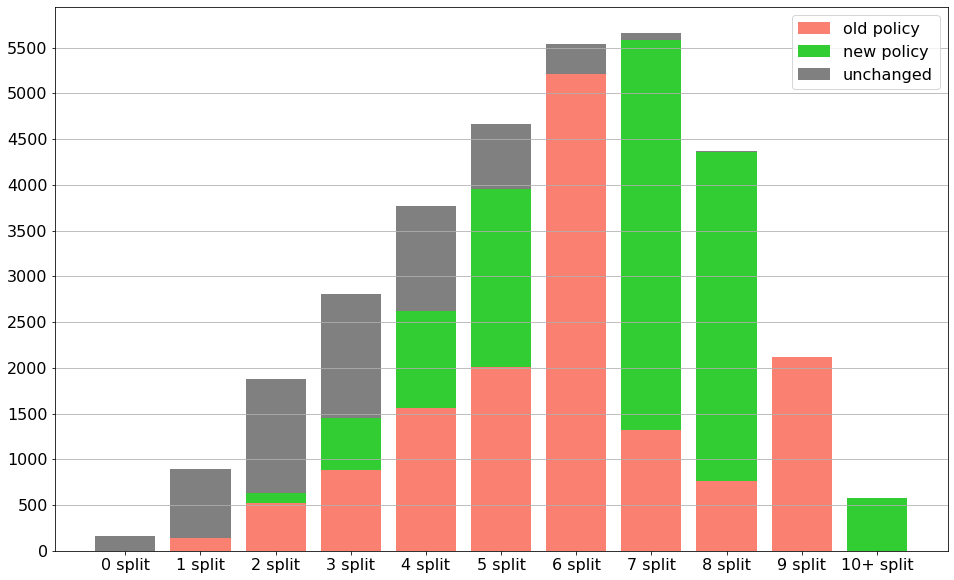}

   \caption{Comparison of the scope of influence of the old and new segmentation alignment policy. The height represents the number of videos in each group. The green part of the bar indicates the part where the new segmentation alignment strategy whereas the red part represents videos that are still using the original strategy.}
   \label{fig:coverage}
\end{figure}

In the new strategy, we tested the most suitable segmentation interval for the above 23 groups of videos, and adjust the interval between the two split spots respectively. The coverage of the newly revised strategy is shown in \cref{fig:coverage}. You can see that all videos in our target (10s, 10 times) group are affected by the new strategy.

After the adjustment on the alignment policy, our target group has 32.99\% of videos with a higher F1 score and only 17.88\% has a lower one, which met our expectations. The results show that the new version of the shift strategy can improve the validation set F1 score by about 0.03\%, and just making adjustments on the 576 videos in the target (10s, 10 times) group only can make an improvement of 0.023\% already. Applying this new strategy to the test set, we get an improvement of 0.04\% on the test F1 score as well.

\subsection{Experimental Results}
The improvements each method brings to the model on the test set are shown in \cref{tab:example}. Eventually, we achieved 86.03\% on the F1-score on the Kinetics-GEBD test set, which is 0.09\% higher than our model last year.

\section{Future Works}
\label{sec:future work}
In the previous section, we mentioned that our model tends to predict more segmentations than the ground truth. We also considered doing dynamic thresholds by setting different thresholds for videos in the 23 groups to control the number of segmentations. It turns out that finding the most suitable threshold for each group through grid search can indeed improve the F1 score on the valid set. However, on the test set, the new segmentation alignment strategy can still guarantee a stable improvement, while the improvement brought by the dynamic threshold is basically negligible.

According to our analysis, the alignment strategy is formulated based on the inherent rules and data distribution of the data set, and our adjustment is also closely related to the number of segments, so it is generally applicable to both the validation and test results. Therefore adjusting these strategies will not cause overfitting. For the dynamic threshold strategy, the threshold selection of the model is more related to the score of the model output. Adjusting the threshold according to the apparent relationships between duration and number of cut points may lead to overfitting.

For future work, we consider training different models for videos with different numbers of segmentations. For example, videos with more and less than 0.55 cuts per second can be divided into two groups to train two models respectively. Thus we can encourage different models to have different sensitivities, rather than simply applying different thresholds to the output of the same model.

\section{Conclusion}
\label{sec:conclusion}

In this report, we introduced our solution for the GEBD tasks. We tried several strategies to improve our model performance based on the same network structure we adopted in the LOVEU-GEBD challenge in 2022 though some strategies did not work. The focal loss and a new segmentation alignment strategy improved our model performance further. There is still much work we can do for GEBD tasks in the future such as we explained in \cref{sec:future work}.

{\small
\bibliographystyle{ieee_fullname}
\bibliography{Paper}
}

\end{document}